\documentclass{article}

\usepackage{PRIMEarxiv}

\usepackage[utf8]{inputenc} 
\usepackage[T1]{fontenc}    
\usepackage{hyperref}       
\usepackage{url}            
\usepackage{booktabs}       
\usepackage{amsfonts}       
\usepackage{nicefrac}       
\usepackage{microtype}      
\usepackage{lipsum}
\usepackage{fancyhdr}       
\usepackage{graphicx}       
\graphicspath{{media/}}     

\usepackage{multirow}
\usepackage{booktabs}
\usepackage{amsmath}
\usepackage{tabularx}

\title{Investigating Neural Machine Translation for Low-Resource Languages: Using Bavarian as a Case Study\thanks{Preprint accepted at SIGUL 2024}}

\author{
 Wan-Hua Her \\
  Information Science\\
  University of Regensburg\\
  Germany \\
  \texttt{wan-hua.her@stud.uni-regensburg.de} \\
   \And
 Udo Kruschwitz \\
  Information Science\\
  University of Regensburg\\
  Germany \\
  \texttt{udo.kruschwitz@ur.de} \\
}

\begin{document}
\maketitle

\begin{abstract}
Machine Translation has made impressive progress in recent years offering close to human-level performance on many languages, but studies have primarily focused on high-resource languages with broad online presence and resources. With the help of growing Large Language Models, more and more low-resource languages achieve better results through the presence of other languages. However, studies have shown that not all low-resource languages can benefit from multilingual systems, especially those with insufficient training and evaluation data. In this paper, we revisit state-of-the-art Neural Machine Translation techniques to develop automatic translation systems between German and Bavarian. We investigate conditions of low-resource languages such as data scarcity and parameter sensitivity and focus on refined solutions that combat low-resource difficulties and creative solutions such as harnessing language similarity. Our experiment entails applying Back-translation and Transfer Learning to automatically generate more training data and achieve higher translation performance. We demonstrate noisiness in the data and present our approach to carry out text preprocessing extensively. Evaluation was conducted using combined metrics: BLEU, chrF and TER. Statistical significance results with Bonferroni correction show surprisingly high baseline systems, and that Back-translation leads to significant improvement. Furthermore, we present a qualitative analysis of translation errors and system limitations.
\end{abstract}

\keywords{Neural Machine Translation, Low-resource Languages, Back-translation, Bavarian, German}

\section{Introduction}

Neural Machine Translation (NMT) has progressed so far to  reach human-level performance on some languages \cite{lample_phrase-based_2018} and has become one of the most prominent approaches within the research area of Machine Translation (MT). Its easy-to-adapt architecture has achieved impressive performance and high accuracy. Promising methods that fall under NMT include Transfer Learning \cite{zhang_two_2021, zoph_transfer_2016}, pre-trained language models \cite{ahmed-etal-2023-enhancing, clinchant-etal-2019-use}, and multilingual models \cite{huang-etal-2023-knowledge, mueller_analysis_2020, aharoni-etal-2019-massively, dabre_exploiting_2019} etc.

However, existing NMT resources focus overwhelmingly on high-resource languages, which dominate a great portion of contents on the Internet and Social Media. Low-resource languages are often spoken by minorities with minimal online presence and insufficient amount of resources to achieve comparable NMT results \cite{maillard-etal-2023-small, feldman_neural_2020}, but they might even have a very large population of speakers and still be under-resourced (such as Hindi, Bengali and Urdu). Growing interest in low-resource MT is evident through the annually held Conference on Machine Translation (WMT). In 2021, WMT featured tasks to promote MT in low-resource scenarios by exploring similarity and multilinguality \cite{akhbardeh_findings_2021}. Among all tasks, the objective of the Very Low Resource Supervised Machine Translation task \cite{libovicky_findings_2021} focused on Transfer Learning between German and Upper Sorbian. The task examined effects of utilizing similar languages and results show that combining Transfer Learning and data augmentation can successfully exploit language similarity during training.

We introduce our experiment to develop bidirectional state-of-the-art NMT systems for German and Bavarian, a classic high-resource to/from low-resource language pair. Inspired by WMT21, our experiment explores the generalizability of Back-translation and Transfer Learning from the highest-ranking approach from \cite{knowles-larkin-2021-nrc}. Our approach covers the following: First, a simple Transformer \cite{vaswani_2017_attention} is trained as the baseline. Secondly, we use the base model for Back-translation and take the extended corpus to train our second model. Lastly, we experiment with Transfer Learning \cite{zoph_transfer_2016} by introducing German-French as the parent model. For evaluation we opt for a combination of three metrics: BLEU \cite{papineni-etal-2002-bleu}, chrF \cite{popovic-2015-chrf} and TER \cite{Snover_2006_TER}. Recent studies have argued that using BLEU as a single metric neglects the complexity of different linguistic characteristics. Using combined metrics and having various penalization standards may be able to capture translation errors more diversely \cite{kocmi-etal-2021-ship, freitag_bleu_guilty_2020}.

By choosing the language pair Bavarian / German we offer one exemplar for a low-resource language (combined with a high-resource one) that can serve as a reference point for further experimental work applied to other low-resource MT. This will ultimately help addressing the imbalance that still prevails between a handful of well-resourced languages and the many others that are not.
This paper makes the following contributions: 

\begin{itemize}
    \item We offer a systematic evaluation of state-of-the-art NMT approaches for a language pair involving a low-resource language that has attracted little attention so far. This investigation explores both translation from as well as into the low-resource language. We focus on a Transformer baseline against Back-translation and a Transfer Learning approach.
    \item To foster reproducibility and replicabilty (which is in the very spirit of SIGUL, LREC and COLING) we make all code available via a GitHub project repository\footnote{\url{https://github.com/whher/nmt-de-bar}}.

\end{itemize}

\section{Related Work}

\subsection{Low-Resource Languages}

The challenges of low-resource languages can be very diverse, hence difficult to define in simple words. 

For a start, even though large web-crawled data such as OPUS \cite{tiedemann-2012-parallel} has resulted in automatically generated parallel corpora for many minor languages, the quality of the data has been reported to be noisy. Examples include the Bantu (Niger-Congo) languages, where parallel data exists, but often too inconsistent to generate desirable MT performance and reproducible benchmarks \cite{reid_afromt_2021}. Misalignments and mistranslations have also been reported while working with multilingual Indian languages \cite{goyal_efficient_2020}. The rise of Unsupervised NMT \cite{chronopoulou_improving_2021, artetxe_robust_2018, lample_etal_unsupervised_2018} alleviates the need for large amounts of labeled training data. Nonetheless, researchers have noted however strong the supervision during training is, there is an overall dependence on parallel data to support evaluation systems \cite{bender_benderrule_2019, guzman_flores_2019}. We therefore see the problem of these less-studied languages as a problem caused by both the \textit{quantity} and the \textit{quality} of the resources. Without linguistically-trained speakers, parallel data is often curated in an unsupervised fashion and therefore noisy.

Furthermore, there are endangered languages \cite{cieri_selection_2016}, for example, the language Bribri is an extremely low-resource indigenous language which is currently being displaced by English and Spanish \cite{feldman_neural_2020}. Aside from suffering digital inequalities and having insufficient written data, it was more challenging to create standardized representations of Bribri, since lexemes and rules vary from communities of speakers. Another similar study which focused on Alemannic dialects also highlights that dialects do not have uniform spelling rules, and that spelling reflect different regional pronunciations \cite{lambrecht-etal-2022-machine}. This raises a great challenge for MT to decide which variation should be given precedence. These under-resourced languages raise a string of challenges due to long years of absence of standardization, and that digital revitalization is not merely a question of gathering data and training models.

To optimize text processing and its size during training, the most common way is to create a joint vocabulary through Byte Pair Encoding (BPE) \cite{sennrich-etal-2016-neural}. BPE is a highly effective subword segmentation algorithm. It iteratively merges frequent words and creates new subword units from infrequent words. A drawback of this approach is that the model learns patterns of smaller unit composition only by recognizing the infrequent words. To counter this, BPE dropout was introduced by \cite{provilkov_bpe_dropout_2020} to stochastically corrupt the segmentation procedure within BPE.

\subsection{Machine Translation}

\paragraph{Nearest Neighbor Machine Translation}

Non- and semi-parametric methods have been successfully applied to MT tasks in recent years.  \cite{gu_2018_nonparametric} demonstrate a powerful combination of neural networks and non-parametric retrieval mechanisms to improve translation. \textit{k}NN-MT follows the retrieval principle and proposes a more efficient non-parametric translation method, which augments the decoder of a pre-trained NMT model with a nearest neighbor retrieval mechanism, allowing direct access to data store of cached examples \cite{khandelwal_2021_knn_mt}. This approach scales the decoder to an arbitrary amount of examples at test time, particularly strengthening decoder's translation capability. However, the big drawback is high computational cost and low decoding speed due to word-by-word generation. Chunk-based \textit{k}NN-MT \cite{martins-etal-2022-chunk} solves this problem by processing translation in chunks of words instead of passing single tokens through the data store.

\paragraph{Transfer Learning} in MT is often done by training a high-resource language pair and using this parent model to initialize parameters in a child model with low-resource languages. For example, \cite{zoph_transfer_2016} achieved translation improvements for Hansa, Turkish and Uzbek into English by using French-English as a parent model. Experiments from \cite{kocmi_trivial_2018} showed improvements using Transformers \cite{vaswani_2017_attention} to train low-resource languages such as Estonian and Slovak. Their results pointed out key factors for a successful transfer include the size of the parent corpus and sharing the target or source language. For instance, Estonian-English as a child gained up to 2.44 BLEU with Finnish-English as a parent. 


In Dual Transfer \cite{zhang_two_2021}, two parent models are used to initialize one child. Monolingual and parallel parent data were trained separately so that inner layers and embeddings can be transferred separately. Another recent study extends conventional transfer learning by additionally transferring probability distributions from parent to child. The Consistency-based Transfer Learning \cite{li-etal-2022-consisttl} argues that parent prediction distribution is highly informative and can be useful to guide child translation. Their experiment showed that using German-English as a parent can achieve BLEU improvement up to 6.2 for Indonesian-English. Furthermore, the study from \cite{huang-etal-2023-knowledge} investigated a technique to incrementally add new language pairs to a multilingual MT model based on knowledge transfer, without posing the original model at risk for catastrophic forgetting.

\paragraph{Pre-trained Language Models} (PLMs) can be fine-tuned on low-resource languages. For instance, MT quality between Spanish and Quecha was shown to improve by leveraging Spanish-English and Spanish-Finnish PLMs \cite{ahmed-etal-2023-enhancing}, with the latter yielding better results. Furthermore, \cite{imamura-sumita-2019-recycling} combined a BERT \cite{devlin_bert_2019} encoder with a vanilla NMT decoder. Evaluation on low-resource languages like English-Vietnamese show that their two-stage training improves performance significantly compared to simple fine-tuning. XLM extends the features of BERT by using Cross-Lingual Masked Language Modeling \cite{lample_conneau_2019_cross}. It has not only been reported to be beneficial for general unsupervised learning, but also for low-resource supervised MT such as English-Romanian. \cite{gheini_cross-attention_2021} acknowledged the success of PLMs and presented their granulated study of fine-tuning, which showed that cross-attention layers are crucial to continue training downstream tasks and that they are powerful when adapting to new languages.

\subsection{Refined Solutions}
\paragraph{Data Filtering and Normalization}

Translation data for low-resource languages are very difficult to come by and the primary source are often from the Web, making the data noisy and of poor quality \cite{batheja-bhattacharyya-2022-improving}. Extra analysis and text normalization are often required to prevent overfitting. For instance, inaccurate translations, noisy data and a large amount of text-overlap was found in the parallel data for African languages collected from large crowd-sourced platforms \cite{reid_afromt_2021}. Comparative results showed that an English-Zulu model trained with noisy data leads to unreliable results and a reduction of 7 BLEU. Research from \cite{guzman_flores_2019} corroborated this and provided guidelines for removing low-quality translations. They presented translation filtering by way of n-gram models trained on monolingual data and sentence-level char-BLEU score \cite{denoual-lepage-2005-bleu} below 15 or over 90. Another novel filtering approach was proposed by \cite{batheja-bhattacharyya-2022-improving}, where cosine similarity is determined based on available parallel (good quality) data, which is then used as the threshold to filter out pseudo-parallel (noisy) sentences.

\paragraph{Multilinguality}

Previous findings have pointed out that one-to-many models with middle-sized parallel corpora have achieved better results than one-to-one models \cite{dong-etal-2015-multi}. The multilingual model consisting of seven Asian languages developed by \cite{dabre_exploiting_2019} using the Asian Language Treebank \cite{thu-etal-2016-introducing} is a great example. The presence of multiple in-domain aligned languages was argued to have contributed to better learn joint representations, hence leading to intra-language improvements. However, low-resource languages often face the risk of being overfitted in multilingual setups \cite{elbayad-etal-2023-fixing}.
\cite{mueller_analysis_2020} investigated the extent of multilinguality for low-resource languages. Their corpus consists of Bible texts in 1,108 languages, all aligned by verse. Results show that BLEU increase/decrease with respect to the number of training languages is not uniform across languages. Although the 5-language models outperform bilingual baseline models for Turkish and Xhosa, accuracy decrease can be found in Tagalog. The negative correlation between number of languages and translation quality is found to start at 10 languages, and maximal degeneration is observed at 100 languages, where addition of languages does not affect translation fluency anymore. This complication and pattern of degeneration can be explained by \cite{holtzman_2019_text_degeneration}, where text repetition harms the likelihood function during decoding. Furthermore, the errors in sequence modeling are more obvious for multilingual corpora, indicating that increased number of languages leads to increased destructive interference. 

\paragraph{Language Similarity}

Leveraging similarities between low-resource languages has been a growing interest in the MT community and is evident through the Similar Language Translation task (SLT) and Very Low Resource Supervised Machine Translation task at WMT21 \cite{WMT:2021}. Regardless of level of closeness and degree of mutual structures, similarity between languages has shown to have positive interactions with MT quality \cite{adebara-etal-2020-translating}. The goal of using language relatedness is similar to leveraging multilinguality. The major difference is they often do not use English as the pivot language, but translate between closely-related languages.

In the Very Low Resource Supervised Machine Translation task at WMT21 \cite{libovicky_findings_2021} between German and Upper Sorbian, the participants were encouraged to make use of Czech and Polish datasets (languages closely related to Sorbian). Results pointed out the importance of including related languages, and that carefully applying tricks can compensate for using smaller datasets substantially. For example, NoahNMT's \cite{zhang-etal-2021-noahnmt} approach entails a Dual Transfer \cite{zhang_two_2021} model that was initialized using German and Czech monolingual data as a parent model. The NRC-CNRC team's \cite{knowles-larkin-2021-nrc} high-performance was attributed to the combination of minor tricks such as Back-translation \cite{sennrich_backtranslation_2015}, monolingual data selection by way of consine similarity, Moore-Lewis filtering \cite{moore-lewis-2010-intelligent} and BPE dropout \cite{provilkov_bpe_dropout_2020}.

The technique Back-translation is further backed up by the study from \cite{lambrecht-etal-2022-machine}. They investigated the effect on Alemannic dialect translation and experienced significant improvement, suggesting that Back-translation is a highly promising method for low-resource languages.

\section{Methodology}

Motivated by the current findings, we present our experiment to develop bidirectional state-of-the-art NMT systems between German and Bavarian (ISO codes are de and bar respectively) - a language pair consisting of high- and low-resource languages. 
While Bavarian and Upper Sorbian are very different languages, they are both spoken by communities which are geographically located within or near Germany. We expect that applying the NMT methods that were found to be effective as part of  WMT21 might result in similar findings for our setting.

We formulate the following three research questions (applied to the exemplar language pair Bavarian / German):

\begin{itemize}
    \item \textbf{RQ1}: Does translating between similar languages achieve generally higher BLEU scores?
    \item \textbf{RQ2}: How well does Back-translation perform for (bidirectional) German-Bavarian?
    \item \textbf{RQ3}: Does cross-lingual transfer lead to improved results for German-Bavarian? More specifically, does the child model profit from related parent languages (i.e. German-French)?
\end{itemize}   


\subsection{Data Acquisition}

The Tatoeba Challenge\footnote{\url{https://github.com/Helsinki-nlp/tatoeba-challenge}} \cite{tiedemann-2020-tatoeba} is one of the most active projects advocating low-resource MT. It maintains a leader board to compare submitted MT system performance from the community. To our knowledge, we are the first to conduct MT for German-Bavarian systems. We discovered parallel and monolingual sources on OPUS\footnote{\url{https://opus.nlpl.eu/}} \cite{tiedemann-2012-parallel}, which we used for our experiments. More information about data sources can be found in our repository.


\subsection{Framework}
\label{methodology:framework}
Inspired by the WMT21 Very Low Resource Supervised Machine Translation task \cite{libovicky_findings_2021}, our experiment revisits solutions that have been proven to work effectively with low-resource languages.

\begin{itemize}
    \item First, a simple Transformer \cite{vaswani_2017_attention} model using preprocessed parallel data is trained as the baseline model. 
    \item Secondly, Back-translation is used to generate silver-paired parallel data to increase corpus size. 
    \item Lastly, we experiment with Transfer Learning \cite{zoph_transfer_2016} by introducing German-French as the parent model. 
\end{itemize}

For evaluation, we opt for an ensemble of automated MT metrics consisting of BLEU, chrF and TER for our systems. This is backed up by recent argumentation from \cite{kocmi-etal-2021-ship} and \cite{freitag_bleu_guilty_2020}, which states that multiple metrics instead of a single metric can diversify the evaluation based on different linguistic characteristics. This approach is a growing trend and has also been adopted by WMT21. Moreover, the study from \cite{lambrecht-etal-2022-machine} pointed out BLEU is insufficient in word matching due to ununified orthography.

\section{Implementation}

\paragraph{Data Preparation}

In total we found 99.7K parallel sentences between Bavarian and German on OPUS (details can be found in our repository). After extensive preprocessing, the corpus size was reduced to 42K. To conduct data augmentation for the second system, we downloaded an extra 258K of German and 295K Bavarian monolingual text, mainly from Wikipedia and Wikinews. For German-French, we collected a total size of 184K of parallel data from Tatoeba and WikiMedia, which was reduced to 165K after preprocessing. We argue that the amount of in-domain data could contribute positively to Transfer Learning. Text preprocessing removes special symbols and noisy annotation, as proposed in previous studies \cite{knowles-larkin-2021-nrc, goyal_efficient_2020}.

In addition to conventional text preprocessing, we took two further measures to de-noise the data. The additional measures entail check and remove misaligned texts by way of cosine similarity between source and target languages and smart sentence truncation. Based on the knowledge that Bavarian and German share common script and that many morphemes are alike, cosine similarity is a great way to support misalignment removal. We assume that a low cosine correlation indicates a low relevance in context between source and target. Following exploratory experiments, we set the correlation threshold at 0.48 and treat anything that falls below 0.48 as misalignment and remove this. We leave a systematic investigation into this aspect as future work.

Our consideration for smart truncation comes from the long-tailed distribution of sentence lengths (outliers span up to 8000). Having long sentences in the corpus therefore poses potential threat that could damage MT performance \cite{koehn_six_2017}. However, if all longer sequences were simply removed, we might lose a significant amount of precious parallel data. Therefore, we implemented smart truncation to deal with longer sequences in the parallel corpus. The truncation is set at the sequence length of 90. 

\paragraph{Cross Validation} In low-resource MT training, it is important to implement Cross Validation (CV) to ensure robust predictive performance and address problems like overfitting. In this case, where the training corpus is small, CV can provide insights on the variability. We opt for 5-fold CV to compare training results. After text preprocessing, the cleaned text are randomly shuffled and split into 5 chunks. The subsets are then concatenated respectively before training. For our baseline systems, 4 of 5 iterations have the subset size of 33813 for training and 8453 for test. The last iteration has the size of 33812 and 8454 respectively.

\paragraph{System Implementation} of all three systems is carried out as explained in Section \ref{methodology:framework}. We utilized the MT development toolkit Sockeye \cite{domhan-etal-2020-sockeye} for BPE encoding, model training and evaluation.

\paragraph{Statistical Significance} 

For statistical significance analysis, our experimental setup needs to take the multiple comparison problem into account. When testing multiple hypotheses simultaneously, the increased number of statistical inferences leads to increased probability of inexact inferences and Type I errors, making the conventional \textit{p} threshold of 0.05 less reliable. This is a well-known problem, e.g.  in the Genome- and Public Health-related research \cite{aickin_gensler_1996_multi_test, noble_2009_multi_test}. 

Methods that counteract multiple testing generally adjust \(\alpha\) so that the chance of observing inaccurate significant result is reduced. The Bonferroni correction is the simplest (and fairly conservative) approach to cut off the \(\alpha\) value. Bonferroni corrects the \(\alpha\) by considering the set of \textit{n} comparisons, causing the \(\alpha\) threshold to become \(\alpha/\textit{n}\). With the Bonferroni correction, the \textit{p}-value is set to 0.017 as opposed to 0.05.

\section{Evaluation}

\subsection{Metrics}

Despite the popularity of BLEU, recent studies from \cite{kocmi-etal-2021-ship} and \cite{freitag_etal_errors_2021} questioned the phenomenon of using BLEU as a single metric, especially in low-resource scenarios, where language structures and scripts are complex and different from many high-resource languages. For example, the meta evaluation on Indian languages by \cite{sai-b-etal-2023-indicmt} reported higher human judgement correlation using COMET \cite{rei-etal-2020-comet} as opposed to BLEU. The limitation of BLEU also lies in the strong dependence on reference translation, whose quality can be highly unstable, especially when data is noisy. Issues such as translationese and poor reference diversity \cite{freitag_bleu_guilty_2020} might also jeopardize the entire evaluation. We therefore include chrF and TER for a more diverse evaluation. ChrF is language-independent and has been reported to better capture complex morpho-syntactic structures in MT evaluation \cite{popovic-2015-chrf}. TER (Translation Error Rate) quantifies the amount of edit operations it takes to change the system output to match the reference translation \cite{Snover_2006_TER}. This intuitive technique avoids knowledge-intensive calculations and focuses on matching hypothesis with reference. The main advantage of TER as opposed to BLEU is the lower penalty for phrasal shifts. TER has also been reported to correlate highly with human judgement and has been implemented in recent WMT tasks \cite{akhbardeh_findings_2021, mathur-etal-2020-results}.




\subsection{System 1: Baseline}

Despite the lack of sufficient amount of parallel data, baseline models in both translation directions exceed 60 BLEU (see Table \ref{tab:overview_system_results}). For bar-de baseline, BLEU scores have an average of 66, chrF has an average of 78 and TER 33. We want to point out little variation between the folds - indicating that the results are robust. However, we observe relatively lower scores on the opposite direction, namely an average of 61 BLEU, 74 chrF and 36 TER. Variation are also small for the de-bar base systems.

\subsection{System 2: Back-translation}

Back-translation (BT) was applied to the best performing baseline folds with monolingual data. Significant improvements can be observed in all three metrics for bar-de, whereas de-bar systems show subtle increase. In contrast to baseline systems, we observe a systematic increase of standard deviation. Where SD was between 0.3 and 0.6 for base systems, 0.7 to 2.2 SD was found in back-translated systems. 

\subsection{System 3: Transfer Learning}

In contrast to surprisingly high baselines, both parent models perform similarly moderate, the fr-de model scored 29 BLEU, 52 chrF and 65 TER, whereas the de-fr parent reached 30 BLEU, 53 chrF and 65 TER. Given the fact that the German-French corpus size is significantly bigger than the German-Bavarian corpus, we had expected better performance of the parent models. However, our results are comparable with available German-French models on Hugging Face, for instance the one from Helsinki-NLP\footnote{\url{https://huggingface.co/Helsinki-NLP/opus-mt-fr-de}}.

Despite the parents' BLEU scores are only a half of our baseline models, Transfer Learning improves children's performance considerably. For bar-de, the best system has 54 BLEU, 71 chrF and 42 TER, which is an increase of 25 BLEU and 19 chrF and decrease of 23 TER. For de-bar, the best model scored 51 BLEU, 65 chrF and 43 TER, which has a performance leap of 21 BLEU, 12 chrF and 22 TER from parent. We note that Transfer Learning improved translation capacity from parent to child with an enhancement of more than 20 BLEU. This corroborates with the recent studies on the use of Transfer Learning for low-resource languages. However, these improvement cannot compare with the very high baseline systems and their back-translated extensions. 

\begin{table}[t!]
\small
    \centering
    \begin{tabular}{llcccc}
         & \textbf{Model} & \textbf{BLEU} & \textbf{chrF} & \textbf{TER} \\
         \toprule
        \multirow{3}{4em}{bar-de} & Baseline & 66.0 & 78.1 & 32.7 \\
         & Back-translated & 73.4 & 82.5 & 25.0 \\
         & Transferred & 53.9 & 70.5 & 41.9 \\
         \midrule
        \multirow{3}{4em}{de-bar} & Baseline & 61.2 & 74.4 & 36.2 \\
         & Back-translated & 63.4 & 76.3 & 31.9 \\
         & Transferred & 48.2 & 63.9 & 44.4 \\
         \bottomrule
    \end{tabular}
    \caption{Overview of best performing models from each system}
    \label{tab:overview_system_results}
\end{table}

\subsection{Statistical Analysis}

Two-tailed pairwise t-tests were conducted on all pairs with Bonferroni correction (\textit{p} threshold is 0.017). Test statistics are shown in Tables \ref{tab:ttest_bar-de} and \ref{tab:ttest_de-bar}. For bar-de models, the BLEU results from baseline (M = 65.7, SD = 0.2) and BT (M = 70.5, SD = 2) indicate that Back-translation leads to significant improvement, \textit{t} = -4.89, \textit{p} = 0.0036. BT also performs significantly better than transferred systems (M = 52.8, SD = 0.7), \textit{t} = 17.25, \textit{p} < 0.0. Further statistics from the metrics chrF and TER corroborate these findings.

For de-bar models, the tendency is similar. ChrF results show a positive enhancement from baseline (M = 74.1, SD = 0.4) to BT (M = 75.5, SD = 0.7), \textit{t} = -3.84, \textit{p} = 0.149. The improvement of BT over transferred systems (M = 64.2, SD = 0.6) is significant as well. TER statistics also verify these findings. Interestingly, while chrF and TER successfully rejects the null hypothesis between baseline and BT performance, BLEU does the opposite. We argue that the results are nevertheless significant based on chrF and TER, and consider this disagreement between metrics as an occurrence derived from linguistically-different perspectives and computations.

\begin{table*}[t]
\small
    \centering
    \begin{tabular}{llllllc}
    \toprule
        \textbf{Metric} & \textbf{Group 1} & \textbf{Group 2} & \textbf{\textit{t}} & \textbf{\textit{p}} & \textbf{\textit{p} (corr.)} & \textbf{Reject \(\text{H}_{0}\)} \\
    \midrule
        \multirow{3}{3em}{BLEU} & Baseline & BT & -4.89 & 0.0012 & 0.0036 & True \\ 
         & Baseline & Transfer & 37.86 & 0.0 & 0.0 & True \\ 
         & BT & Transfer & 17.25 & 0.0 & 0.0 & True \\
    \midrule
        \multirow{3}{3em}{chrF} & Baseline & BT & -5.83 & 0.0004 & 0.0012 & True \\ 
         & Baseline & Transfer & 20.65 & 0.0 & 0.0 & True \\ 
         & BT & Transfer & 19.82 & 0.0 & 0.0 & True \\
    \midrule
        \multirow{3}{3em}{TER} & Baseline & BT & 6.1 & 0.0003 & 0.0009 & True \\ 
         & Baseline & Transfer & -19.29 & 0.0 & 0.0 & True \\ 
         & BT & Transfer & -16.2 & 0.0 & 0.0 & True \\
    \bottomrule
    \end{tabular}
    \caption{Results of t-test with Bonferroni correction for bar-de systems.}
    \label{tab:ttest_bar-de}
\end{table*}

\begin{table*}[t]
\small
    \centering
    \begin{tabular}{llllllc}
    \toprule
        \textbf{Metric} & \textbf{Group 1} & \textbf{Group 2} & \textbf{\textit{t}} & \textbf{\textit{p}} & \textbf{\textit{p} (corr.)} & \textbf{Reject \(\text{H}_{0}\)} \\ 
    \midrule
        \multirow{3}{3em}{BLEU} & Baseline & BT & -2.85 & 0.0214 & 0.0641 & False \\ 
         & Baseline & Transfer & 29.58 & 0.0 & 0.0 & True \\ 
         & BT & Transfer & 22.04 & 0.0 & 0.0 & True \\
    \midrule
        \multirow{3}{3em}{chrF} & Baseline & BT & -3.84 & 0.005 & 0.0149 & True \\ 
         & Baseline & Transfer & 30.12 & 0.0 & 0.0 & True \\ 
         & BT & Transfer & 26.28 & 0.0 & 0.0 & True \\
    \midrule
        \multirow{3}{3em}{TER} & Baseline & BT & 5.02 & 0.001 & 0.0031 & True \\ 
         & Baseline & Transfer & -23.74 & 0.0 & 0.0 & True \\ 
         & BT & Transfer & -15.91 & 0.0 & 0.0 & True \\
    \bottomrule
    \end{tabular}
    \caption{Results of t-test with Bonferroni correction for de-bar systems.}
    \label{tab:ttest_de-bar}
\end{table*}

\subsection{Qualitative Analysis}

We argue that the surprisingly high baseline results come from the similarity of the source and target languages. This corresponds to findings from \cite{adebara-etal-2020-translating} that language relatedness contributes positively to MT quality. The analysis of \cite{goyal_efficient_2020}'s multilingual NMT on Indo-Aryan languages lists linguistic characteristics such as word-order construction, degree of inflection, amount of similar word root, meaning and conjunct verbs as the key drivers for improving training. Our experiments corroborate these argumentation, thus answering \textbf{RQ1}.


The significant improvement from Back-translation, which can be seen with all metrics, aligns well with previous findings. Especially in the submitted systems for WMT21 Very Low Resource Supervised MT between Upper Sorbian and German by \cite{knowles-larkin-2021-nrc}, Back-translation boosted the training corpus size and contributed to performance increase. However, we are aware of its limits. For instance, the augmented text includes many errors, which were inherited from the baseline systems. This issue of \textit{Translationese} \cite{graham-etal-2020-statistical} is widely discussed, especially in the context of using silver-paired data for MT. In our case, we have opted for a smaller amount of augmented data, with the aim to reduce Translationese as much as possible while still allowing model improvement. We therefore answer \textbf{RQ2} that Back-translation contributes positively.

Regarding \textbf{RQ3}, we point out that while Transfer Learning did improve performance from parent to child, its final performance was not sufficient to exceed the other two systems.

We note that our results are similar to the ones from the German - Upper Sorbian translation task from WMT21. Our baseline and back-translated models have an accuracy range between 60 to 73 BLEU and 74 to 82 chrF, comparable with the final scores from the German - Upper Sorbian task. However, it is interesting to note that their chrF scores are substantially higher than ours (by 10), while our BLEU scores are similar. This brings us back to the notion that all metrics work linguistically different and these variations reflect through different languages.

Furthermore, a common finding can be observed between our experimental results and the WMT21 experiments we comapre against, namely the result discrepancy between high-to-low and low-to-high directions. In our study, de-bar is ca. 10 BLEU and 10 chrF behind bar-de. Similarly but not as extreme, Upper Sorbian - German also performs better than its high-to-low counter direction. This performance gap on the same corpus but different translation directions raises attention, with possible reasons due to the multiple orthographic standards and sub-dialects in our case.

\begin{table*}[t]
\small
    \centering
    \begin{tabular}{l|c|l}
    \toprule
        \textbf{German Input} & \textbf{System} & \textbf{Bavarian Output} \\
    \midrule
        \multirow{2}{12em}{sie hat heute abend im restaurant fisch bestellt.} & Base & se hod heit abend im restaurant fisch bestöid. \\
         & BT & se hod heid obend im restaurant fisch bestejd. \\
    \bottomrule
    \end{tabular}
    \caption{Examples of German to Bavarian translation.}
    \label{tab:translation_examples}
\end{table*}

Table \ref{tab:translation_examples} depicts two translation examples. We translate the German phrase ``Sie hat heute Abend im Restaurant Fisch bestellt" (English meaning ``she ordered fish in the restaurant tonight.") into Bavarian using all of our systems. We observe that while Base and BT outputs look similar, their differences could come from various sub-dialects in the corpus. For instance, the term ``heute" was translated into ``heit" and ``heid", with only the last consonant different. However, in the Germanic linguistics, these consonants ``t" and ``d" differ themselves in voice. The linguistic notion of \textit{Fortis and Lenis}\footnote{\url{https://en.wikipedia.org/wiki/Fortis_and_lenis}} differentiates oral pressure that is given to these consonants. Thus, we suspect these differences come from various dialects.

\section{Conclusion}

In this paper, we presented experimental work in Neural Machine Translation with the aim to  push forward our understanding of how to best address the gap between a handful of  well-resourced languages and the long tail of languages for which no sufficient resources are available. 
More specifically, we focused on methods and case studies that have shown promising results for languages with limited resources. We conceptualized the problems of noisy data and data shortage by way of recent studies. We revisited creative solutions designed to combat these challenges such as Back-translation, multilingual training and language relatedness.
Our own low-resource implementation utilized data augmentation and cross-lingual transfer on German and Bavarian. We report our steps to preprocess the corpus and carry out training for three bidirectional systems. 5-fold cross validation was carried out on each system to compare robustness. We opted for a combined metric system using BLEU, chrF and TER to evaluate translation from different perspectives. For multiple hypothesis testing, pairwise t-tests with Bonferroni correction were conducted to test for statistical significance. 
Results show that translation between similar languages performs generally better and that augmented data contribute positively. However, even though cross-lingual transfer showed huge improvement from parent to child, it was not able to exceed baseline and back-translated models. We recognize that Transfer Learning is an effective approach for low-resource languages, but note that in our study language similarity played a more important role.
To support reproducibility and replicability all  code is made available via GitHub.

\section{Limitations}

The Bavarian orthography has been a known problem for decades, as it is mostly a spoken language and has not been properly standardized. For example, the word 'Bavarian' alone can be written in two ways: Boarisch or Bairisch. The investigation by \cite{zehetner_1978_bairisch} illustrates that there are multiple Bavarian orthographic conventions. From a computational perspective, the issue is ``deciding which representation should be given precedence", as stated in the Bribri case study by \cite{feldman_neural_2020}. Overcoming dialectal variations is also a problem of politics that can carry on for years. In light of the findings by \cite{mager-etal-2023-ethical}, we would add that the automated translation of Bavarian should - like other under-sourced languages - be carefully planned with ethical considerations, and that purely using web-scraped data to deploy translation systems might neglect the concerns of speakers.
Another challenge lies in multiple sub-dialects. This phenomenon can be observed in our corpus, which is mined from the Bavarian Wikipedia, where articles are written in different regional dialects. We argue that these sub-dialects in the parallel corpus lead to translation confusion, resulting in translation outputs which consist of mixed accents. Nevertheless, should there be a more refined and organized corpus of a particular sub-dialect, our systems can serve as baselines for fine-tuning.
Another, more general limitation is the fact that throughout our work we conducted purely technical evaluations. The strength of such an experimental setup is that it can be reproduced and offers objective results. However, it is clearly necessary to involve native speakers  to gain more insights into the quality of any translation process. We mitigated against the problem by choosing not just a single evaluation metric (such as BLEU), but no matter how many different metrics are chosen they are no substitute for user studies.

\section{Future Work}

Following our findings and the limitations stated above, we propose further research directions to inspire future work: First, the curation of a more refined and organized parallel corpus for modern German-Bavarian to help establish a high quality benchmark for training and evaluation. An example to achieve this is through recruiting native speakers in both Bavarian and German who have an adequate amount of linguistic knowledge. This annotation could include not only translation of parallel sentences, but also the sub-dialects or Bavarian regional variations the speakers associate themselves with. This human-annotated dataset could furthermore be split into two parts, one for training and another for evaluation.

Additionally, identification of dialects would be an approach to counter translation confusion and mixed accents. This could help unify and isolate non-standardized languages or dialects. As mentioned in the previous section, a great way to start modelling sub-dialect detection is to automatically analyze the Wikipedia articles with their corresponding sub-dialects. This would greatly reduce the training corpus size, but additional measures to increase the corpus size could be taken, such as acquiring diverse datasets (i.e. open-source subtitles of Bavarian TV-programs or historical documents). More generally, we see our work as a reference benchmark for future work -- be it to explore the same language pair further or other work into the general problem of low-resource language translation efforts. 

\section{Ethical Considerations}

Ethical concerns  arise whenever natural language is being sampled and used to train machine learning systems. For this experimental work we used existing test collections and other freely accessible data. All the experiments are conducted within the ethical framework imposed on us by our institution. In this context we did not identify a specific ethical issue. 

However, it is clear that once any automated  translation system is on its way to be deployed that care must be taken to (a) train it on \textit{representative} samples, (b) mitigate against common biases, and (c) make sure no personal information is included in the training data. If trained on social media data there is also a risk that toxic content might surface. Care must be taken to take these issues seriously (rather than treating this as a box-ticking exercise), but we would argue that there are no ethical concerns arising from this work that have not already been identified previously. 

\section{Acknowledgment}

We would like to thank the anonymous reviewers for their constructive feedback.

\bibliographystyle{unsrt}  
\bibliography{templateArxiv}

\end{document}